\documentclass[11pt,draftcls,onecolumn,journal]{IEEEtran}
\usepackage{amsmath,amsfonts}
\usepackage{algorithmic}
\usepackage{algorithm}
\usepackage{array}
\usepackage{textcomp}
\usepackage{stfloats}
\usepackage{url}
\usepackage{verbatim}
\usepackage{graphicx}
\usepackage{cite}
\usepackage{hyperref}
\usepackage{booktabs}
\usepackage{graphicx}
\usepackage{subfigure}

\usepackage{chngcntr} 
\counterwithin{paragraph}{subsection} 

\usepackage{cleveref}
\NewDocumentCommand{\CoTutor}{}
{
  \unskip\textsc{C\small oT\small u\small t\small o\small r}\unskip
}

\NewDocumentCommand{\CAMP}{}
{
  \unskip\textsc{C\small A\small M\small P}\unskip
}

\begin{document}

\title{Future-Proofing Programmers: Optimal Knowledge Tracing for AI-Assisted Personalized Education}

\author{Yuchen Wang, Pei-Duo Yu, Chee Wei Tan
\thanks{Yuchen Wang and Chee Wei Tan are with the College of Computing and Data Science, Nanyang Technological University, Singapore. Pei-Duo Yu is with the Department of Applied Mathematics, Chung Yuan Christian University, Taiwan.}}

\markboth{IEEE Signal Processing Magazine,~Vol.~XX, No.~XX, June~2024}%
{Aviyente \MakeLowercase{\textit{et al.}}: Author Guidelines for Special Issue Articles of IEEE SPM}

\maketitle

\section*{}
\label{sec:abstract}
Learning to learn is becoming a science, driven by the convergence of knowledge tracing, signal processing, and generative AI to model student learning states and optimize education. We propose \CoTutor{}, an AI-driven model that enhances Bayesian Knowledge Tracing with signal processing techniques to improve student progress modeling and deliver adaptive feedback and strategies. Deployed as an AI copilot, \CoTutor{} combines generative AI with adaptive learning technology. In university trials, it has demonstrated measurable improvements in learning outcomes while outperforming conventional educational tools. Our results highlight its potential for AI-driven personalization, scalability, and future opportunities for advancing privacy and ethical considerations in educational technology. Inspired by Richard Hamming's vision of computer-aided `learning to learn,' \CoTutor{} applies convex optimization and signal processing to automate and scale up learning analytics, while reserving pedagogical judgment for humans, ensuring AI facilitates the process of knowledge tracing while enabling learners to {\it uncover} new insights. 

\section*{Introduction}
\IEEEPARstart{T}{he} rapid advancement of generative AI, particularly large language models (LLMs), has opened new frontiers in education, offering unprecedented opportunities for personalized learning and adaptive allocation of instructional resources (including lecture materials, assignments, and supplementary content). Tools like GitHub Copilot, built on models such as OpenAI's Codex \cite{chen2021evaluatinglargelanguagemodels}, have demonstrated the potential of AI to assist in programming tasks by generating context-aware code snippets. However, the broader challenge in AI-assisted education lies in understanding and modeling student learning states to provide timely and individualized feedback. While Bayesian Knowledge Tracing (BKT) has a long history in educational systems, many traditional implementations were not designed to support the real-time adaptivity now feasible in the GenAI era. More recent systems in the GenAI era, such as Khanmigo \cite{khanmigo2023} and GitHub Copilot in the Classroom \cite{copiloteducation2024}, demonstrate the growing potential of large language models in supporting student learning. These systems leverage LLMs to offer contextualized help, scaffolded explanations, and Socratic prompts, and have been piloted in programming, math, and science domains. They typically rely on end-to-end generative inference or retrieval-augmented prompts and do not maintain an explicit, interpretable model of the student's learning trajectory.

\begin{figure}[t]
    \centering
\includegraphics[width=0.8\textwidth]{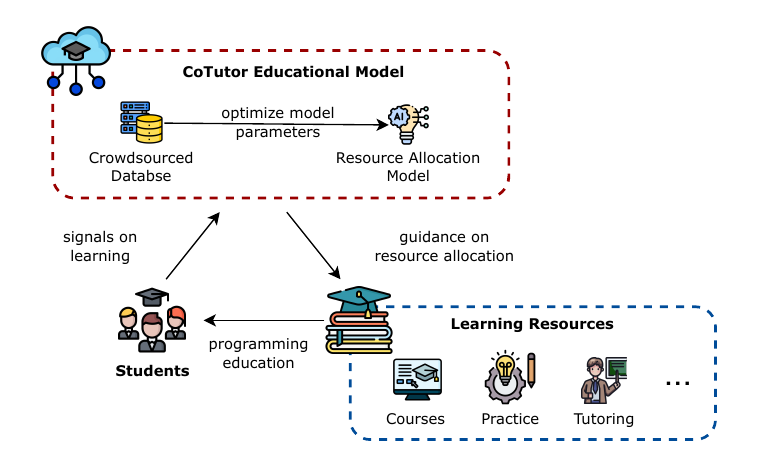}
    \caption{\textbf{Architecture of the \CoTutor{} Programming Education Model.} The model serves as a bridge between students and educational resources, guiding optimal resource allocation and instructional strategies through the processing of student learning signals.}
    \label{fig:loop}
    \vspace{-5mm}
\end{figure}

This paper introduces \CoTutor{}, a generative AI model that integrates knowledge tracing, convex optimization, and signal processing to enhance the delivery of personalized education. \CoTutor{} leverages real-time student interaction data and learning signals to continuously update its understanding of student progress. As DeepMind CEO Demis Hassabis puts it, youth should become ``AI ninjas" by mastering new tools, while grounding themselves in fundamentals and learning how to learn. This assistive role resonates with Richard Hamming's vision for computer-aided instruction \cite{hamming_chap22}, where machines take over routine tasks—such as drilling, grading, and content delivery—allowing human educators to concentrate on fostering creativity, critical thinking, and deeper conceptual understanding—a dichotomy encapsulated by his maxim on learning to learn: ‘{\it What you learn from others you can use to follow; what you learn for yourself you can use to lead.}' As shown in Figure~\ref{fig:loop}, \CoTutor{} operates as a generative AI chatbot that accompanies students throughout their learning journey, adapting to performance in real time and delivering personalized feedback. Its core knowledge tracing mechanism builds on probabilistic reasoning to estimate skill mastery \cite{knowledge_tracing_survey}. 

A key innovation of \CoTutor{} lies in its explicit formulation of knowledge tracing and feedback adaptation as a convex optimization problem, which brings multiple advantages over purely neural or generative approaches. First, convex optimization offers interpretability by making the learning signal and decision boundaries transparent to instructors and knowledge tracing system designers. Second, the convex structure enables globally optimal tracking of student state updates over time, which supports debugging and pedagogical oversight. Third, convex models require significantly fewer computational resources during inference compared to autoregressive LLMs, enabling efficient deployment at scale. Unifying probabilistic reasoning, signal processing, and convex optimization positions \CoTutor{} as a practical educational tool and a research-friendly framework to study transparent and ethical AI in education. To evaluate \CoTutor{}, we conducted a two-part study: (1) objective benchmarking on educational datasets and (2) a controlled classroom study measuring student learning outcomes, engagement, and retention. The results indicate that \CoTutor{} outperforms previous methods and enhances the effectiveness of student learning in real world settings.

\begin{figure*}[t]
    \centering
    \begin{subfigure}
        \centering
        \includegraphics[width=0.45\textwidth]{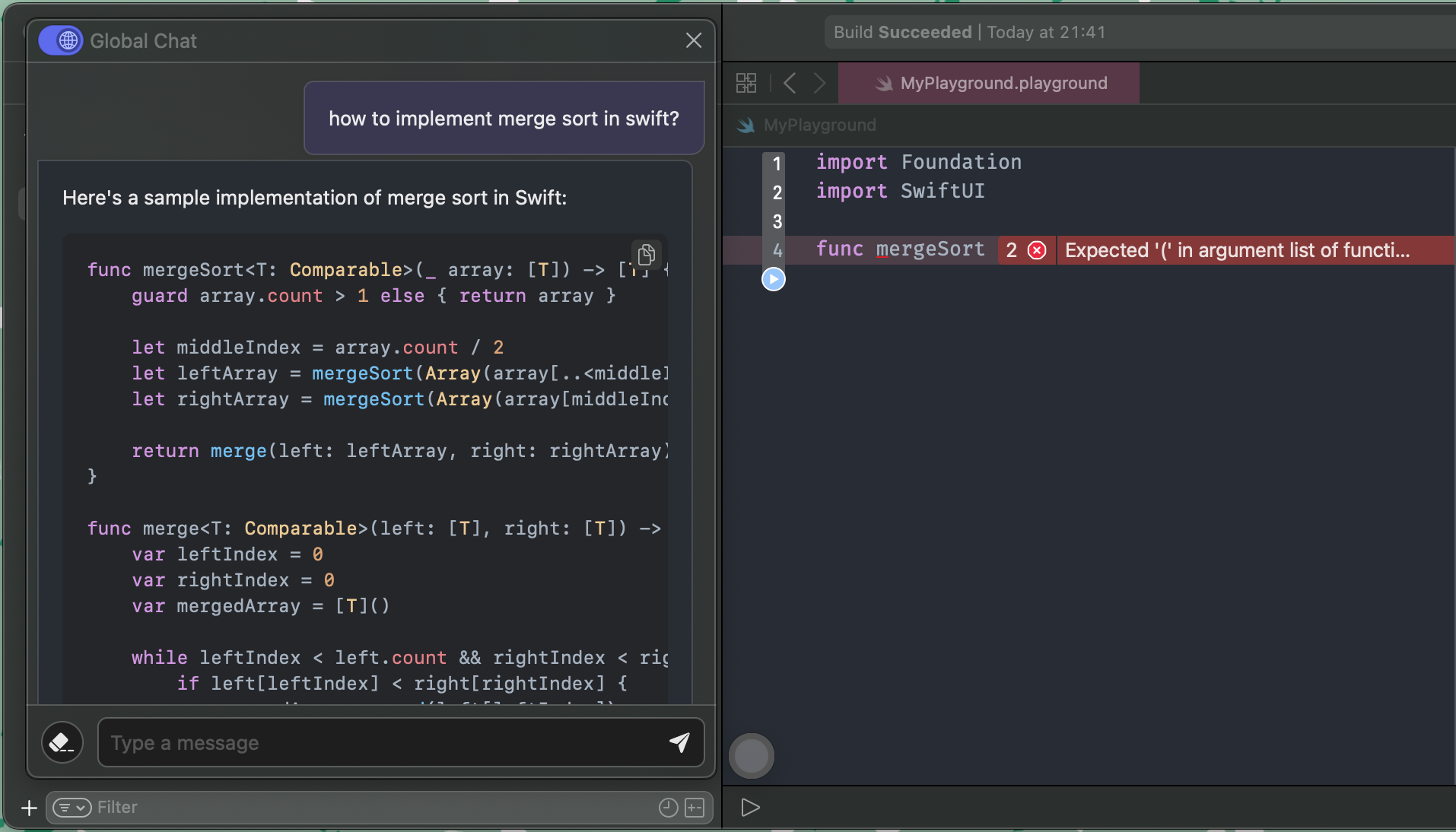}
    \end{subfigure}%
    ~ 
    \begin{subfigure}
        \centering
        \includegraphics[width=0.45\textwidth]{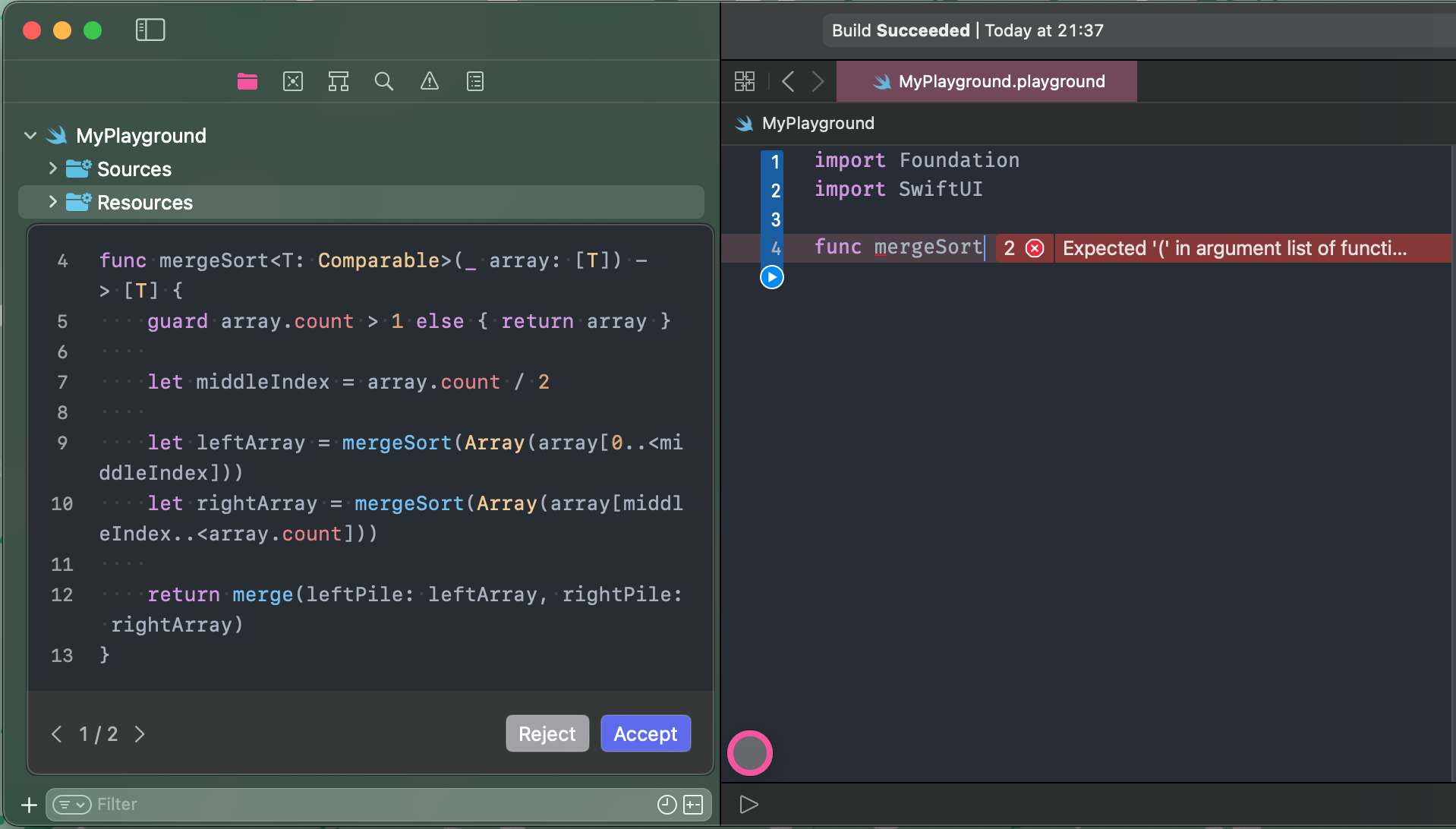}
    \end{subfigure}
    \caption{\textbf{User interface of Copilot for Xcode}, which supports generative programming tasks such as code auto-completion, chatbot-based Q\&A, and fix suggestions.}
    \label{fig:copilot_for_xcode}
    \vspace{-5mm}
\end{figure*}

\section*{Related Work}
This section reviews prior research in AI-assisted education, knowledge tracing, AI-driven educational platforms, and signal processing in education. We highlight key advancements, limitations, and insights that motivate the design of \CoTutor{}.

\subsection*{Generative AI-Assisted Education}
AI-assisted education has seen significant advancements in recent years, driven by the integration of LLMs and adaptive learning systems. Tools like GitHub Copilot, built on OpenAI's Codex \cite{chen2021evaluatinglargelanguagemodels}, have demonstrated the potential of AI to assist in programming tasks by generating context-aware code snippets. These tools not only reduce development time but also help learners understand programming concepts through real-time feedback \cite{chatgpt_pilot_study}. Recent work has made significant strides in improving the accuracy and reliability of AI-assisted tools \cite{wang2025testing, wong2024aligning}. Building on this, Retrieval Augmented Generation (RAG) techniques have been leveraged to equip LLMs with deeper contextual understanding. An example is the Contextual Augmented Multi-Model Programming (\CAMP) framework by \cite{wang2025contextual}. The user interaction surface of \CAMP, illustrated in Figure \ref{fig:copilot_for_xcode} of the implementation as \textit{Copilot for Xcode} \cite{tan2023copilot}, showcases its practical application. While these advancements underscore the transformative potential of AI in education, they often fall short in dynamically adapting to individual student needs in real time, a gap our work seeks to address.

\subsection*{Knowledge Tracing Techniques in Education}
Knowledge Tracing (KT) is a fundamental technique for modeling student learning dynamics and predicting students' future performance. Traditional methods, such as BKT, use probabilistic reasoning to estimate the likelihood that a student has mastered a specific skill based on their performance on related tasks across multiple learning sessions \cite{yudel_BKT}. Deep Knowledge Tracing (DKT) addresses some limitations of BKT in handling complex, real-time interactions and diverse learning trajectories by using recurrent neural networks to model temporal dependencies in student performance, but it often requires large datasets and lacks interpretability \cite{piech2015deep}. Recent advancements in KT include Control Knowledge Tracing \cite{loong2024control}, which applies control-theoretic principles to model the dynamic evolution of learning, and Comprehension Factor Analysis \cite{thaker2019comprehension}, which incorporates external factors like reading behavior and practice to improve predictions. These methods highlight the importance of modeling both learning gains and forgetting rates, as memory decay follows a power-law function \cite{rubin1996one}. \CoTutor{} extends KT by integrating real-time signal processing and generative AI, enabling it to adapt to dynamic learning environments and provide actionable insights for resource allocation.

\subsection*{AI-Driven Educational Platforms}
AI-driven educational platforms have transformed how students learn and interact with educational content. MOOCs and platforms like Khan Academy have integrated AI to personalize learning experiences through knowledge tracing and deep learning techniques \cite{gervet2020deep}. These platforms track student understanding over time, allowing for tailored instruction that addresses individual strengths and weaknesses. Within computing and programming education, these platforms are particularly transformative in facilitating instruction in programming fundamentals, algorithm design, and computational thinking through structured problem-solving approaches. Online Judge (OJ) systems, such as \textit{Codeforces}, \textit{LeetCode}, and \textit{AtCoder}, are widely used in programming education to automatically evaluate code submissions. They provide immediate verdicts (e.g., Accepted, Wrong Answer) based on predefined test cases, enabling learners to refine solutions and improve problem-solving skills. These systems generate rich data on student programming behavior, which can be leveraged for knowledge tracing models \cite{online_code_execution_system}. Next-generation OJ systems are increasingly incorporating AI-driven features, such as error pattern classification, personalized hints, and adaptive problem difficulty. \CoTutor{} exemplifies this trend by leveraging generative AI to provide interactive, personalized feedback and resource allocation, while leveraging the rich data generated for knowledge tracing.

\subsection*{Personalized Learning Foundations}
Personalized education aims to tailor learning experiences to individual students' needs, abilities, and progress \cite{personalized_learning_survey}. Traditional one-size-fits-all teaching methods often fail to address varying learning paces and comprehension levels, leading to disengagement or knowledge gaps. Advances in AI and data mining have enabled adaptive learning systems that dynamically adjust instructional content, pacing, and feedback \cite{pl_next_era}. The foundations of personalized education were first established through heuristic-based pedagogical frameworks which emphasized competency-based progression \cite{bloom}. The emergence of digital learning environments enabled the shift toward data-driven adaptation that used rule-based cognitive models to deliver step-by-step problem-solving guidance \cite{cognitive_tutor}. Meanwhile, educational recommender systems began applying collaborative filtering techniques from e-commerce to personalize resource suggestions \cite{education_recommendation}. While these systems demonstrated early success, they struggled to capture the dynamic nature of learning, particularly in modeling how knowledge states evolve across different skills and contexts. This limitation was addressed through advances in probabilistic modeling and deep learning to capture complex temporal dependencies in learning behavior \cite{yudel_BKT, piech2015deep}, followed by transformer-based architectures and multi-task learning frameworks \cite{saint+, qdkt} that enable more nuanced modeling of both knowledge acquisition and forgetting patterns.

To further address the critical challenge of accurately diagnosing a student's knowledge state and recommending appropriate resources, recent work has explored hybrid models that combine knowledge tracing with reinforcement learning to optimize pedagogical strategies \cite{rl1}. Contemporary approaches like deep reinforcement learning via Adaptive Policy Transfer have further enhanced these methods by incorporating meta-learning for cross-domain policy adaptation. These innovations have significantly advanced the field, though challenges remain in handling sparse data scenarios and maintaining pedagogical interpretability.

\subsection*{Convex Optimization in Educational Resource Allocation}
Traditional methods of educational resource allocation in education often rely on heuristic rules or static progression schedules \cite{bloom}. Such approaches may ignore trade-offs across student heterogeneity, engagement cost, and intervention effectiveness. To the best of our knowledge, there is no previous work focusing on AI-driven educational resource management catered to real-time student needs in interactive scenarios like e-learning platforms. Convex optimization has proven effective in other domains with constrained resource distribution problems \cite{boyd_book}. 
It provides a principled alternative by formulating objectives like:
\begin{equation}
\min_{\mathbf{w}} \sum_i c_i w_i \quad \text{s.t.} \quad \sum_i f_i(w_i) \geq \gamma,\quad 0 \leq w_i \leq 1,
\end{equation}
where $w_i$ represents the intervention weight for objective $i$, $c_i$ the associated cost, and $f_i(w_i)$ the predicted gain. It has the potential to solve educational resource allocation problems catered to individual students' needs, yet its current application in education remains underexplored. \CoTutor{} integrates convex optimization with live knowledge tracing and signal feedback, enabling adaptive scheduling that balances learning impact and system constraints.

\subsection*{Educational Signal Collection and Processing}
Signal processing has emerged as a critical procedure in the educational process, supporting the collection, interpretation, and analysis of diverse student learning signals. Some traditional educational platforms integrate multiple signal sources—including Learning Management Systems (LMS) activity, demographic background, and academic performance—to compute predictive indicators of student risk levels and send personalized alerts that have been shown to improve learning outcomes and retention \cite{signal_purdue}. In digital learning environments such as MOOCs and LMS-based platforms,  signals are often derived from log data (including session duration, clickstream logs, sequence patterns). These are typically pre-processed into derived features—such as temporal patterns, error types, and engagement metrics—for downstream statistical modeling and adaptive interventions \cite{gervet2020deep}. Blended learning implementations further incorporate attendance records, in-class quiz performance, and daily self-reported surveys to enrich the signal space \cite{blended}. \CoTutor{} integrates signal processing to analyze real-time student feedback and engagement data, enabling it to dynamically refine its understanding of student learning states and optimize educational interventions.

\section*{\CoTutor{}: AI-Assisted Educational Model} \label{sec:edusig-defn}

\begin{figure}[t]
    \centering
    \includegraphics[width=0.8\linewidth]{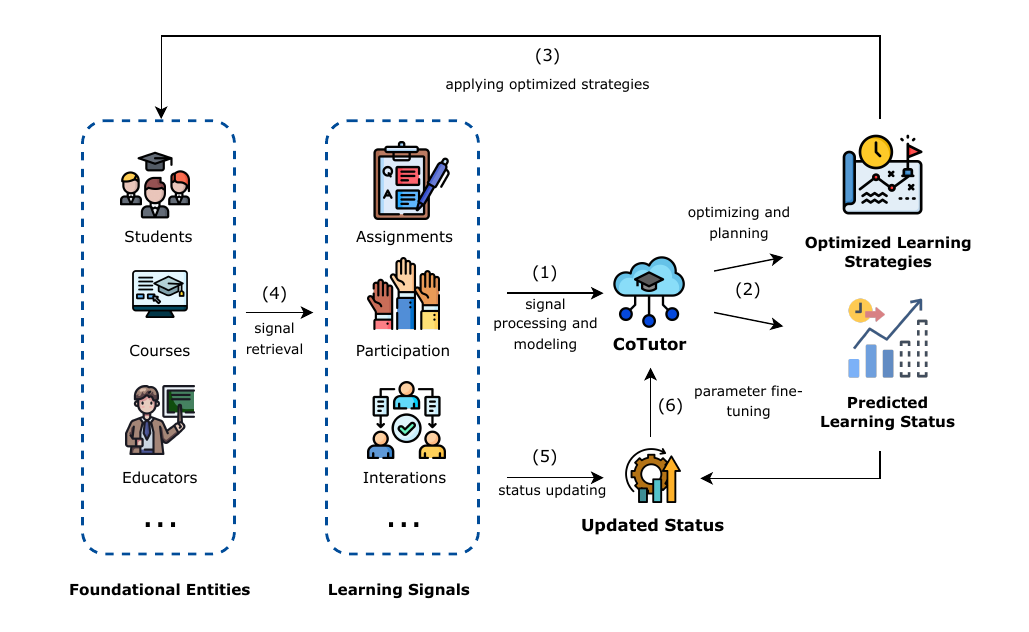}
    \caption{\textbf{The iterative flywheel mechanism of \CoTutor{}.} Multichannel learning signals are collected from student interactions and processed to model students' knowledge states. Based on predicted learning outcomes, the system formulates and applies optimized strategies to refine educational interventions. Knowledge states and parameters are continuously updated to ensure adaptive and personalized learning.}
    \label{fig:flywheel}
    \vspace{-5mm}
\end{figure}

\subsection*{\CoTutor{}: Iterative Signal-Driven Learning Optimization}
We propose \CoTutor{}, an AI-assisted education model that unifies signal processing with probabilistic modeling to optimize personalized learning strategies. As shown by Figure~\ref{fig:flywheel}, at its core is an iterative \textbf{flywheel mechanism} that continuously refines student models and learning interventions by integrating multimodal learning signals and feedback loops. The flywheel comprises three interconnected phases:

\begin{algorithm}
\caption{Iterative Algorithm for \CoTutor{} Model Optimization}
\label{alg:iterative}
\begin{algorithmic}[1]
\STATE \textbf{Input:} Student signal stream $S = \{(t_i, \mathbf{z}_i, y_i)\}$, model state $\mathbf{x}$, policy parameters $\boldsymbol{\theta}$
\STATE \textbf{Parameters:} Smoothing factor $\alpha$, window size $W$, learning rate $\eta$, regularizer $\lambda$
\STATE \textbf{Initialize:} Prior knowledge state $P(L_0)$, latent state $\mathbf{x} \leftarrow \mathbf{x}_0$, policy $\boldsymbol{\theta} \leftarrow \boldsymbol{\theta}_0$

\FOR{each new signal $(t_i, \mathbf{z}_i, y_i)$}
    \STATE \textbf{Signal Processing:}
    \STATE \quad $\bar{\mathbf{z}}_i \leftarrow \alpha \mathbf{z}_i + (1 - \alpha) \bar{\mathbf{z}}_{i-1}$
    \STATE \quad $\tilde{\mathbf{z}}_i \leftarrow \text{Aggregate}(\bar{\mathbf{z}}_{i-W:i})$
    
    \STATE \textbf{Knowledge State Update:}
    \STATE \quad $P(L_i) \leftarrow \text{BKT\_Update}(P(L_{i-1}), y_i, \tilde{\mathbf{z}}_i)$ \COMMENT{Update mastery prob. using Bayes' rule}
    
    \STATE \textbf{Recommendation:}
    \STATE \quad $r_i \leftarrow \text{Recommend}(\mathbf{x}, P(L_i), \boldsymbol{\theta})$
    
    \STATE \textbf{Model Update:}
    \STATE \quad $\mathbf{g} \leftarrow \nabla_{\boldsymbol{\theta}} \text{Loss}(P(L_i), \tilde{\mathbf{z}}_i, y_i)$ \COMMENT{Compute gradient of the loss function}
    \STATE \quad $\boldsymbol{\theta} \leftarrow \text{Prox}_{\lambda}(\boldsymbol{\theta} - \eta \mathbf{g})$ \COMMENT{Apply proximal operator for constrained/regularized update}
    \STATE \quad $\mathbf{x} \leftarrow \text{UpdateLatentState}(\mathbf{x}, \tilde{\mathbf{z}}_i, P(L_i))$ \COMMENT{Refine internal latent representation}
\ENDFOR

\STATE \textbf{Output:} Updated model state $\mathbf{x}$, policy $\boldsymbol{\theta}$, and recommendation $r_i$
\end{algorithmic}
\end{algorithm}

\begin{enumerate}
    \item \textbf{Educational Signal Taxonomy} – \CoTutor{} collects multi-channel learning signals to model student knowledge states. While signal types vary across use cases (detailed in subsequent sections), they fall into three categories: 

    \begin{itemize}
\item \textit{Submission and grading}: Timestamps, attempts, scores, and metadata for assignments, quizzes, and programming exercises.
\item \textit{Participation logs}: Class attendance, forum interactions, and resource access patterns.
\item \textit{Queries and feedback}: Questions from AI chat interfaces (e.g., Q\&A bots) and post-module feedback surveys.
\end{itemize}
These signals are structured as time-series data to capture temporal dynamics in behavior and performance. 

    \item \textbf{Signal Processing and Knowledge State Estimation} – We followed the following procedure: multimodal learning analytics of data digitalization $\rightarrow$ smoothing and normalization $\rightarrow$ sampling $\rightarrow$ feature embedding $\rightarrow$ multimodal data aggregation, to pre-process the collected data, before feeding the data to the modal for learning status modeling, and student status prediction. This aligns with established multimodal learning analytics methodologies proposed by \cite{signal_to_knowledge}.

    \item \textbf{Adaptive Strategy Planning and Iterative Refinement} – Based on the inferred learning states, \CoTutor{} predicts students' next steps and generates optimized learning strategies. These strategies include recommending tailored educational resources, suggesting personalized learning activities, and supporting educators with adaptive intervention guidance. As students interact with the system, \CoTutor{} continuously updates its internal knowledge model with new learning signals, which improves the accuracy of future predictions and the relevance of subsequent recommendations.

\end{enumerate}

\begin{figure}[t]
\centering
\subfigure[Submission history showing scores and resource usage.]{
\includegraphics[width=4cm]{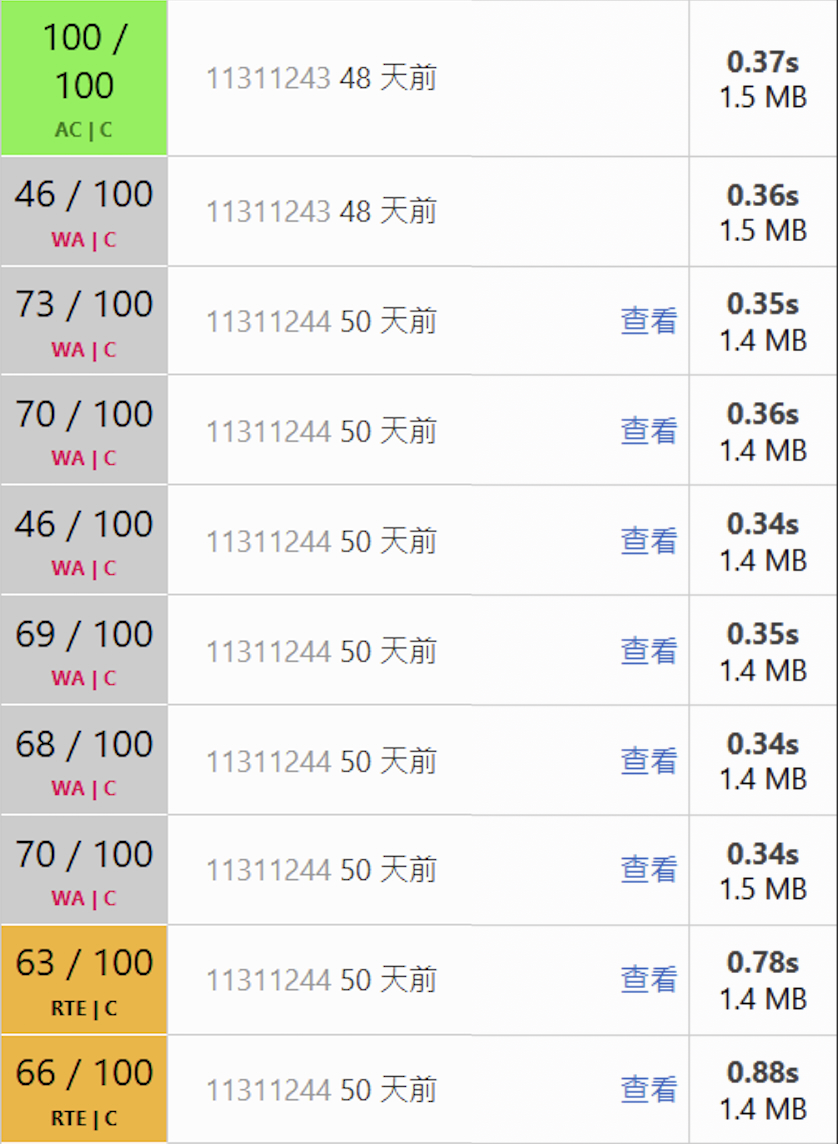}}
\quad
\subfigure[Detailed execution results, including compilation warnings, per-test-case results, and final scores.]{
\centering
\includegraphics[width=6cm]{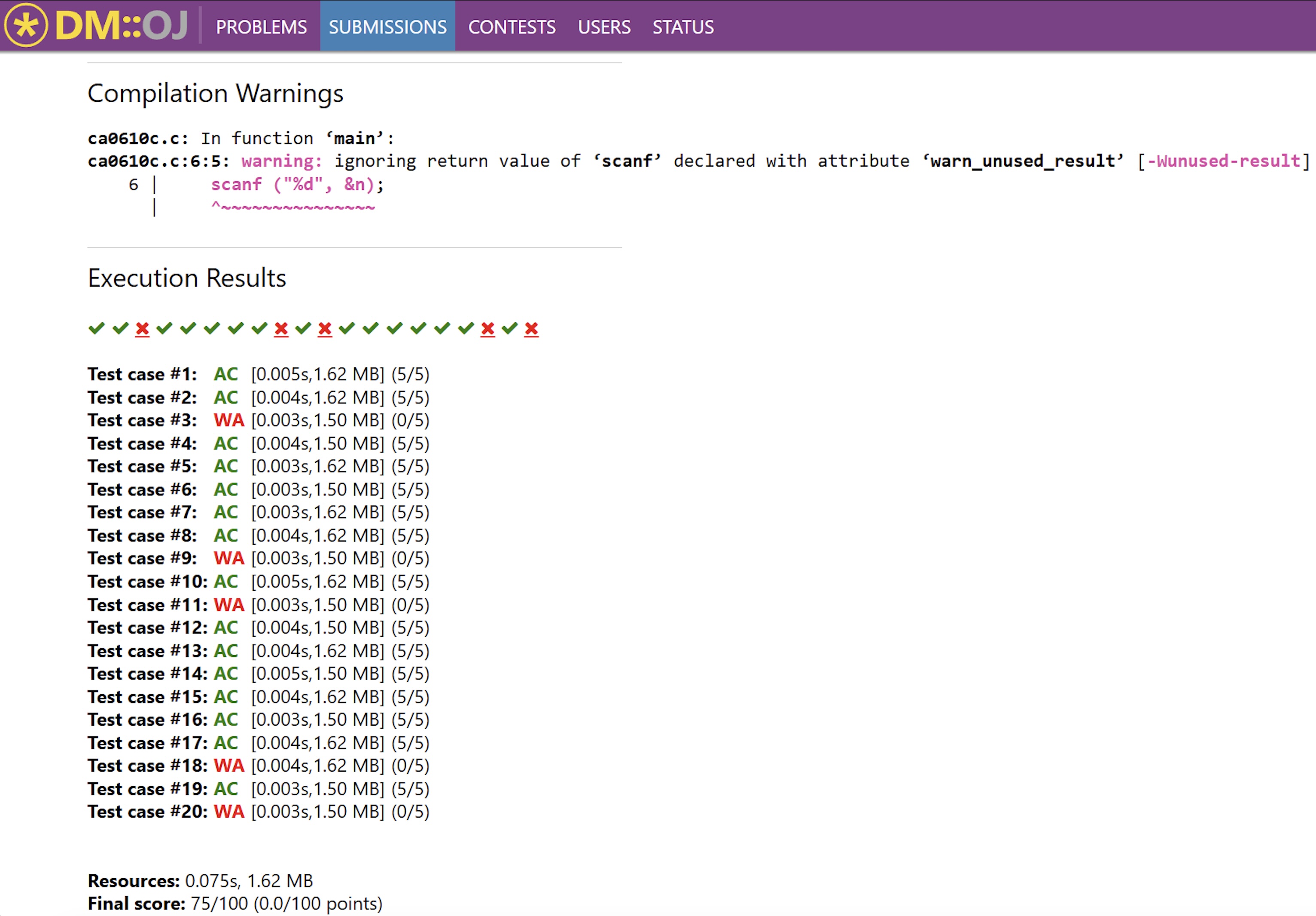}}
\caption{Overview of student submissions on the DMOJ platform in an introductory C programming course.}
\label{fig:dmoj}
\vspace{-5mm}
\end{figure}

We formalize the iterative loop with the following Algorithm \ref{alg:iterative}, where key variables are defined as follows:

\begin{itemize}
    \item $S = \{(t_i, \mathbf{z}_i, y_i)\}$: A stream of learning signals, where $t_i$ is the timestamp, $\mathbf{z}_i$ is a feature vector (e.g., engagement or assessment data), and $y_i$ is the observed performance (e.g., correctness).
    \item $\mathbf{x}$: Latent model state capturing individualized learning representation.
    \item $\boldsymbol{\theta}$: Policy parameters used to generate personalized recommendations.
    \item $\alpha$: Exponential smoothing factor for noise reduction in signal data.
    \item $W$: Aggregation window size for computing smoothed trends over recent interactions.
    \item $\eta$: Learning rate for gradient-based parameter updates.
    \item $\lambda$: Regularization weight to avoid overfitting.
\end{itemize}

This cyclical approach ensures that \CoTutor{} progressively improves its ability to model student progress and personalize learning pathways. The next section presents a practical example of the \textbf{Adaptive Strategy Planning} phase, demonstrating how \CoTutor{} optimally assigns educational resources to meet individual student needs while providing actionable insights for educators.

Our evaluation focuses on two key hypotheses:
\begin{itemize}
    \item \textbf{H1: Improved Knowledge Modeling} – \CoTutor{} achieves superior learning state estimation compared to baseline models by integrating observed signals more effectively.
    \item \textbf{H2: Enhanced Learning Outcomes} – Students assisted by \CoTutor{}, which combines accurate knowledge state prediction and optimized resource recommendations, achieve better learning performance compared to baseline conditions, including traditional methods (no model assistance), BKT, DKT, and code-DKT.
\end{itemize}

The next section presents the implementation of \CoTutor{} within an integrated learning platform and outlines how its components support personalized, scalable, and research-ready educational deployment.

\begin{table}[t]
    \centering
    \caption{Learning signals collected from DMOJ submissions and platform logs.}
    \label{tab:dmoj_data}
    \begin{tabular}{lp{9cm}}
        \toprule
        \textbf{Attribute} & \textbf{Explanation} \\
        \midrule
        Submission identifiers & Unique IDs linking students to specific problems. \\
        Verdict and error type & \texttt{Accepted} (AC), \texttt{Wrong Answer} (WA), \texttt{Runtime Error} (RTE), \texttt{Compilation Error} (CE), etc. \\
        Execution time and memory usage & Quantitative indicators of solution efficiency. \\
        Time of submission & Timestamp for chronological ordering. \\
        Number of attempts before success & Aggregated across consecutive wrong submissions. \\
        Problem difficulty level & Labels such as ``easy,'' ``medium,'' or ``hard,'' or numerical ratings. \\
        \bottomrule
    \end{tabular}
\end{table}

\subsection*{Using Online Judge Data to Estimate \texorpdfstring{$P(L_{j,i})$}{P(L\_{j,i})}}

To implement the proposed \CoTutor{} framework in a real-world educational context, we apply it to an \textbf{Online Judge (OJ)} system used in programming courses. These platforms are widely used to automatically assess code submissions and provide immediate feedback. Their structured environments generate rich data for estimating learning progress, making them ideal for modeling student knowledge. Figure~\ref{fig:dmoj} illustrates student submissions in an introductory C programming course, highlighting key metrics such as scores, resource usage, and per-test-case results.

We represent each student's submission history as a sequence of learning interactions:

\begin{equation}
    D^{(j)} = \{(t_i, \mathbf{z}_i, y_i)\}_{i=1}^{T}.
\end{equation}

Here, $t_i$ is the timestamp, $\mathbf{z}_i$ encodes contextual features, as listed in Table \ref{tab:dmoj_data}, and $y_i \in \{0, 1\}$ denotes binary correctness (\texttt{AC} vs.\ not). We apply exponential smoothing to track recent context:

\begin{equation}
\tilde{\mathbf{z}}_i = \alpha \mathbf{z}_i + (1 - \alpha) \tilde{\mathbf{z}}_{i-1}.
\end{equation}

Using this data, \CoTutor{} updates each student's knowledge state via the BKT framework. Specifically, we define $P(L_{j,i})$ as the probability that a student has mastered activity $j$ in session $i$. This probability is updated based on the student's interaction outcomes and the model's parameters:

\begin{align}
P(L_{j,i}) &= P(L_{j,i-1}\mid \text{obs}) + \left(1 - P(L_{j,i-1}\mid \text{obs})\right)\cdot T_{j}, \\
P(C_{j,i}) &= P(L_{j,i}\mid \text{obs})\cdot(1 - S_{j}) + \left(1 - P(L_{j,i}\mid \text{obs})\right)\cdot G_{j}.
\end{align}

Here:
\begin{itemize}
    \item $T_j$ is the learning rate: probability of transitioning from an unlearned to a learned state.
    \item $S_j$ is the slip rate: probability of answering incorrectly despite mastery.
    \item $G_j$ is the guess rate: probability of answering correctly without mastery.
\end{itemize}

When a student submits code and receives a verdict (correct or incorrect), the model updates their mastery probability $P(L_{j,i})$ via the posterior inference rules of BKT, given a correct response:

\begin{equation}
P(L_{j,i} \mid \text{correct}) 
= 
\frac{
P(L_{j,i}) \,(1 - S_j)
}{
P(L_{j,i}) \,(1 - S_j) 
+ 
(1 - P(L_{j,i}))\, G_j
},
\label{eq:bkt_correct}
\end{equation}

and for an incorrect submission:

\begin{equation}
P(L_{j,i} \mid \text{wrong}) 
= 
\frac{
P(L_{j,i}) \,S_j
}{
P(L_{j,i})\, S_j 
+ 
(1 - P(L_{j,i}))\, (1 - G_j)
}.
\label{eq:bkt_wrong}
\end{equation}

Equations (\ref{eq:bkt_correct}) and (\ref{eq:bkt_wrong}) allow real-time updates to the student's estimated mastery after each submission. Personalized parameters $T_j$, $S_j$, and $G_j$ can be estimated from online quiz data or historical submission records. As students interact with coding exercises through OJ platforms, \CoTutor{} uses this feedback loop to recommend personalized next steps—such as more challenging tasks, review sessions, or targeted interventions.

\begin{figure}[t]
\centering
\subfigure[\textit{My Code Weapon}]{
\includegraphics[width=7cm]{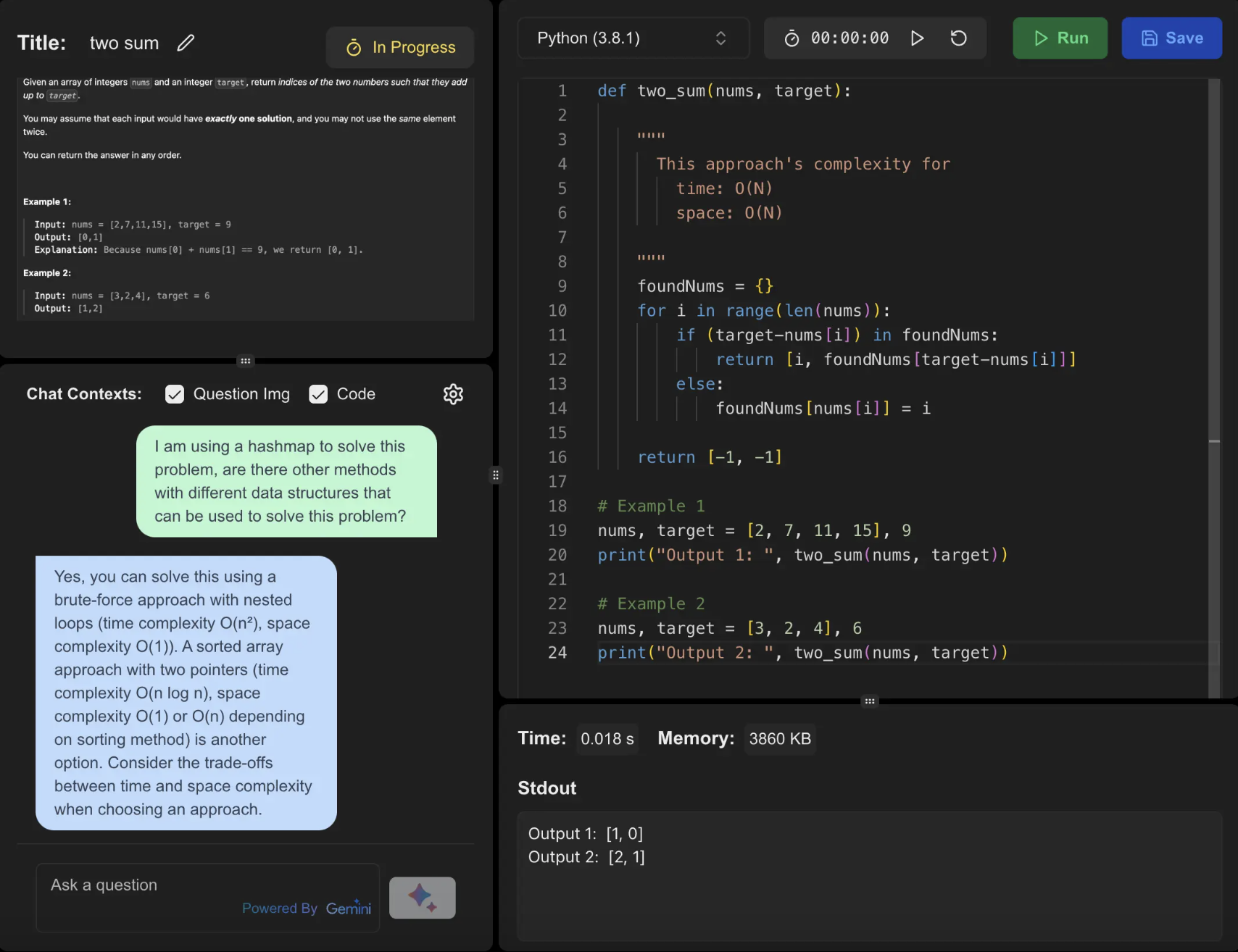}}
\quad
\subfigure[\textit{Nemobot}]{
\centering
\includegraphics[width=8.5cm]{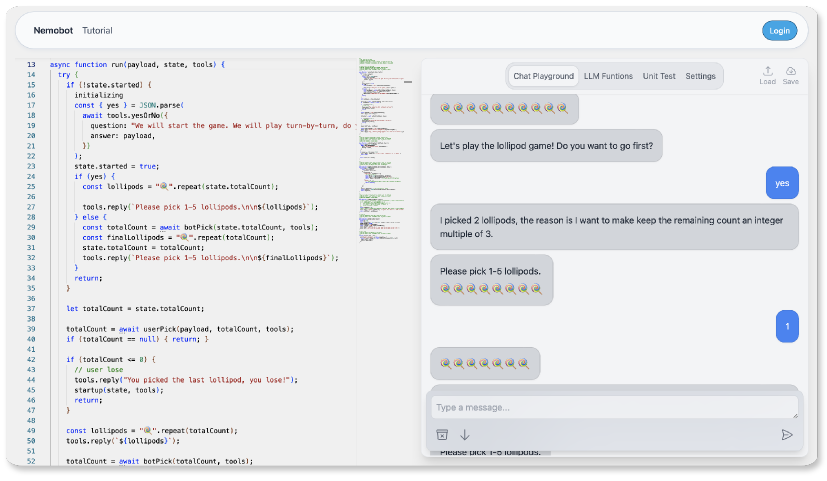}}
\caption{Examples of programming education platforms with AI-powered chatbots showcasing interactive learning functionalities. (a) \textit{My Code Weapon}, a platform for teaching data structures and algorithms to beginner programmers. (b) \textit{Nemobot}, an AI-driven platform for LLM agent programming education using cloud-based large language models and local software tools.}
\label{fig:chatbot}
\vspace{-5mm}
\end{figure}

\subsection*{Interactive Learning Experiences with an AI Copilot}
To enhance student engagement and accessibility, we implemented \CoTutor{} as an interactive AI copilot that seamlessly integrates with online programming education platforms: \textit{My Code Weapon}\footnote{My Code Weapon, freely available at \url{https://www.mycodeweapon.com}, trains junior programmers into code ninjas by connecting code and images for creative problem-solving and exploring algorithmic insights.} and \textit{Nemobot} \cite{wang2024nemobot} (using generative-AI and classroom flipping techniques \cite{tan2021}). The copilot serves as a personalized learning assistant, transforming \CoTutor{}'s outputs into conversational interactions with the following key functionalities, as illustrated by the screenshots from Figure~\ref{fig:chatbot}:

\begin{itemize}
    \item \textbf{Proactive Reminders} – The chatbot sends push notifications to remind students of upcoming deadlines, scheduled activities, and recommended learning tasks.
    \item \textbf{Question Handling} – It supports real-time query resolution by synthesizing responses from multiple sources, including direct educator inputs and generative AI-powered retrieval techniques.
    \item \textbf{Resource Recommendation} – Based on students' learning status, the chatbot provides hyperlinked access to relevant educational materials, ensuring personalized resource allocation.
\end{itemize}

Traditional communication methods—such as navigating online learning platforms, posting in course forums, writing emails to educators, or searching for assigned materials—are streamlined into an intuitive chatbot interface. This approach allows students to proactively engage in their learning process by asking questions, tracking progress, and receiving personalized recommendations in a conversational format. \CoTutor{} adapts content delivery by offering structured, foundational exercises to students with lower mastery and advanced challenges to those with higher mastery, tailoring learning to individual progress for a more effective and engaging experience.

While AI copilots are increasingly used in education, our implementation addresses key limitations of generic systems and distinguishes itself in the following ways:

\begin{enumerate}
    \item \textbf{Methodology-Driven Personalization}: Unlike generic recommendation systems, \CoTutor's chatbot delivers outputs grounded in a principled optimization of student-specific resource allocation, computed from signal-processed knowledge state estimation.
    
    \item \textbf{Continuous Learning Trajectory Integration}: Rather than supporting one-off conversation sessions, \CoTutor{} maintains a persistent model of each student's learning journey and iteratively refines learning path suggestions based on accumulated signals across both assessment and interaction logs.
    
    \item \textbf{Assistive Interface for Experimentation and User Study}: The AI chatbot functions as a delivery and interaction module, enabling more efficient user experiments and study deployment while maintaining methodological rigor in backend inference and planning.

    \item \textbf{Hallucination Mitigation for Educational Reliability}: Given the risk of hallucinations in free-form generative responses, \CoTutor{} standardizes replies where feasible—e.g., linking to lecture timestamps or exercise references—and logs all outputs for instructor review. This design supports transparency and iterative refinement while ensuring reliability in educational contexts.
\end{enumerate}

\section*{Application of \CoTutor{}: A Case Study}
In this section, we formulate the problem of educational resource allocation and the maximization of students' sentiment status as a convex optimization problem. We then demonstrate the application of \CoTutor{} in solving this problem and optimizing educational performance.

\subsection*{Problem Formulation}
Our problem formulation is inspired by the exercise from \cite{boyd_exercise}. We consider a sequence of $n$ education sessions, denoted as $E_1, E_2, \dots, E_n$. Each session $E_i$ consists of $m$ types of educational resources (e.g., theoretical content, applications, experiments) represented as $R_{1,i}, R_{2,i}, \dots, R_{m,i}$. The total resource allocation for each session is constrained as follows:
\begin{equation}
E_i = [R_{1,i}, R_{2,i}, \dots, R_{m,i}], \quad \text{such that} \quad \sum_{j=1}^{m} R_{j,i} = 1, \quad \forall i \in {1, \dots, n}.
\end{equation}
This ensures that the sum of the allocated resources is a normalized unit value per session.

We impose a priority ranking constraint on resource allocation, where certain resources must be allocated proportionally before others. Specifically, given two resource types $a$ and $b$, the following condition must hold up to session $j$:
\begin{equation}
\sum_{i=1}^{j} R_{a,i} \leq  g(\sum_{i=1}^{j} R_{b,i}), \quad \forall j \in {1, \dots, n}.
\end{equation}
This constraint ensures that resource $a$ requires pre-allocation of at least a certain amount of resource $b$. For each student, we define the sentiment status at session $i$ as $s_i$, where:
\begin{equation}
s_i > 0 \text{ (happy)}, \quad s_i = 0 \text{ (neutral)}, \quad s_i < 0 \text{ (unhappy)}.
\end{equation}
The sentiment status is influenced by both the previous session's status and the resources allocated in the current session $i$, modeled as follows:
\begin{equation}
s_i = \theta s_{i-1} + (1 - \theta) f(E_i),
\end{equation}
where $\theta \in [0,1]$ represents the student's emotional volatility, indicating how quickly they react to the content of recent lectures. Here, $f(\cdot)$ represents a hypothesized function of the resources that model the status. In our experimental trials in the next section, we present several models of $f$ for comparison.

Our objective is to determine the optimal allocation of resources $R_{ji}$ for each education session $i$ that maximizes the terminal sentiment status of students at the final session $n$:
 \begin{small}
\begin{align*} 
\text{maximize}   & \;\; \sum\limits_{i=1}^n [\theta s_{i-1} + (1 - \theta) f(E_i)], \label{eq:opt-obj}  \\ 
\text{subject to} & \;\;\; \sum\limits_{j=1}^m R_{j,i} \leq B_i, \;\;  \forall i=1,2,...,n, \\
                  & \;\;\;\sum_{i=1}^{j} R_{a,i} \leq  g(\sum_{i=1}^{j} R_{b,i}), \quad \forall j \in {1, \dots, n},\\
                  & \;\;\;R_{j,i}\geq 0, \;\;\forall i,j,  \\
\text{variables}  & \;\;\; R_{j,i} \;\;  \forall i,j.  \\
\end{align*} 
\end{small}

This problem can be formulated as a convex optimization problem \cite{boyd_book}, subject to the constraints on resource allocation and priority ranking as discussed above, provided the function $g(\cdot)$ is concave and $f(\cdot)$ is chosen such that the objective is concave.

\begin{figure*}[t!]
    \centering
    \begin{subfigure}
        \centering
        \includegraphics[width=0.65\textwidth]{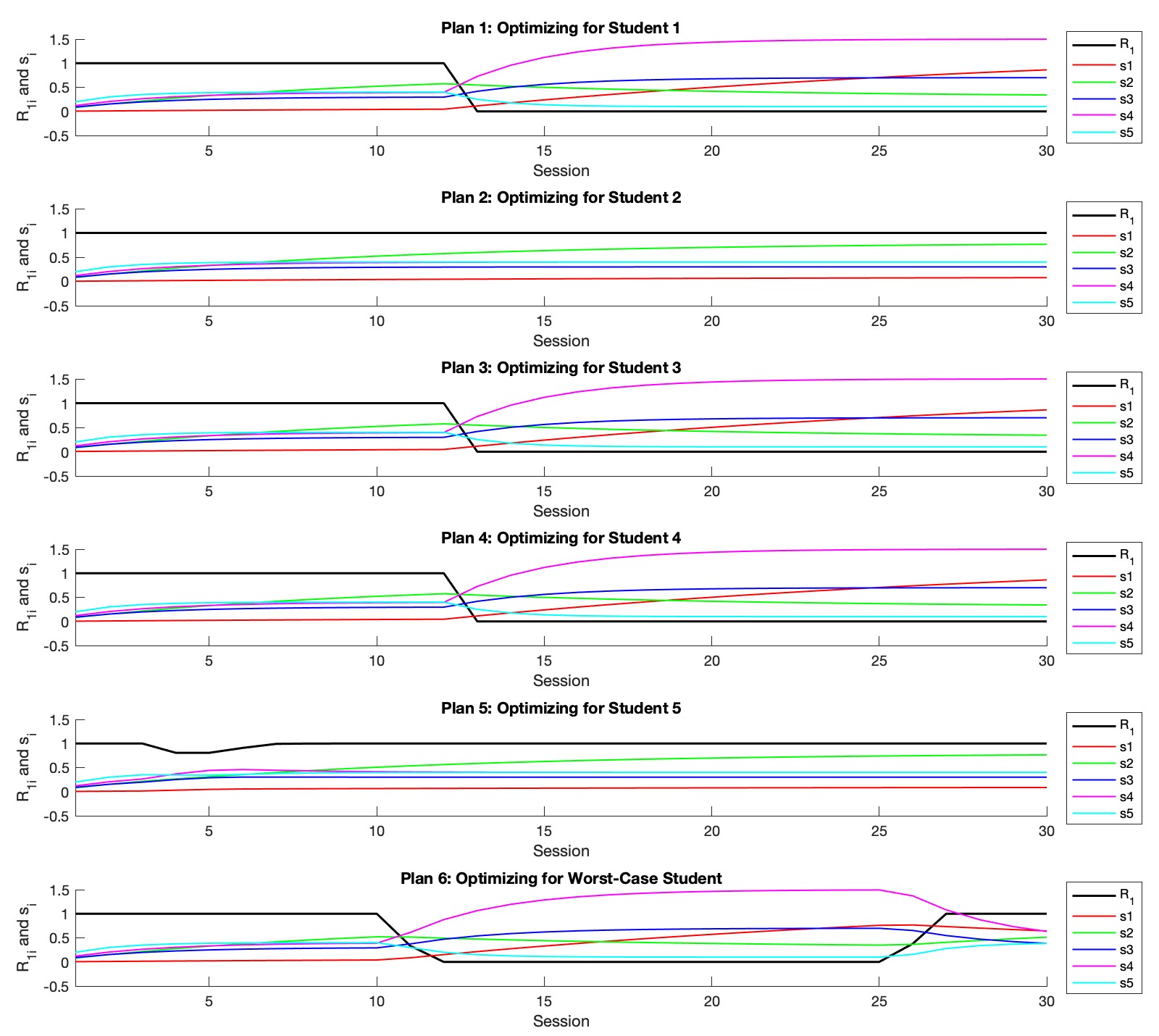}
        \caption{Resource allocation results under linear influence function.}
        \label{fig:allocation_linear}
    \end{subfigure}%
    ~ 
    \begin{subfigure}
        \centering
        \includegraphics[width=0.65\textwidth]{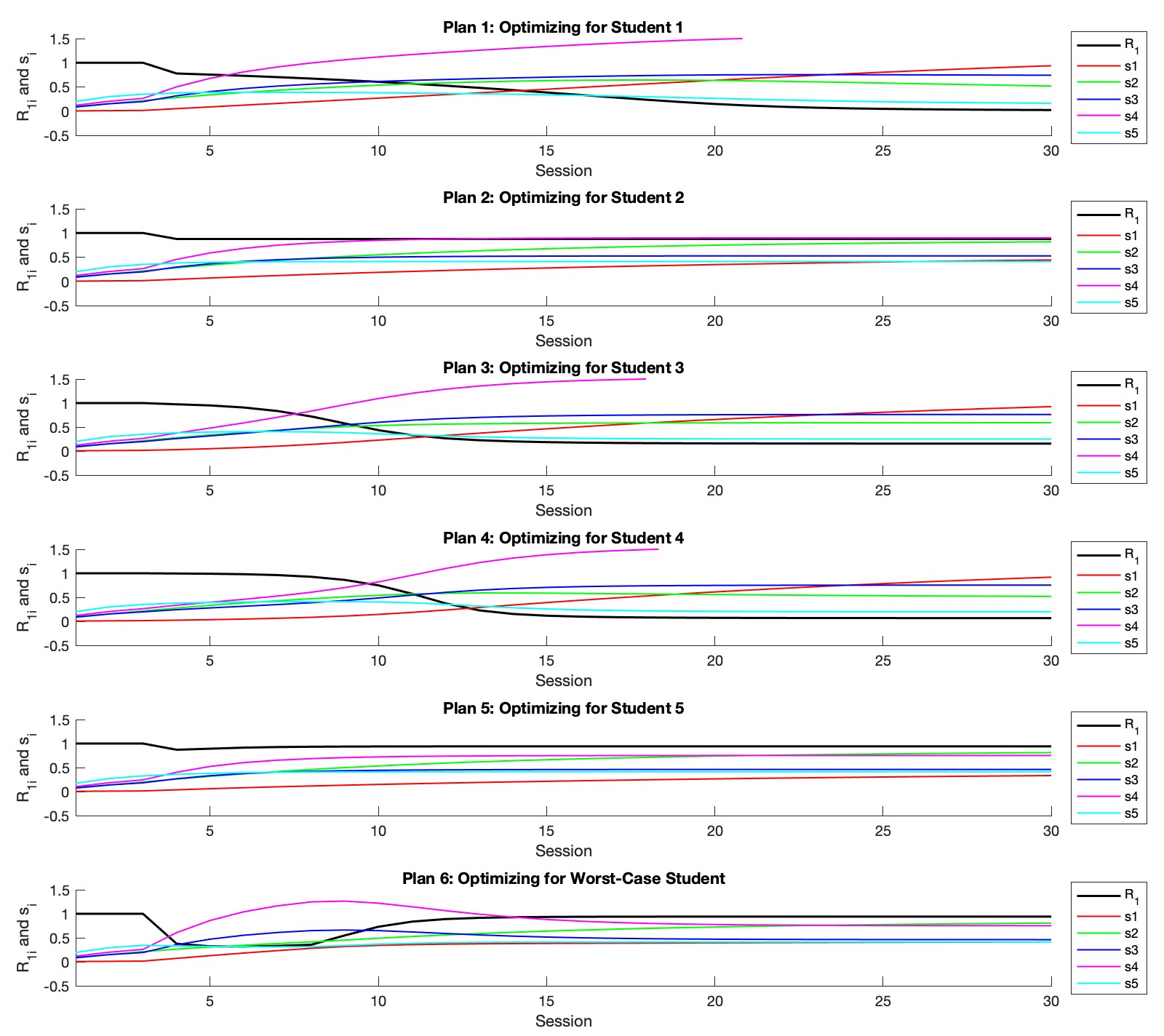}
        \caption{Resource allocation results under power function influence ($k=0.5$).}
        \label{fig:allocation_power}
    \end{subfigure}
\end{figure*}

\subsection*{Solution with \CoTutor{}}
\CoTutor{} optimally allocates resources by solving the convex optimization problem under various influence models. We analyze the performance under different configurations with 30 educational sessions ($n = 30$) and two resource types ($m = 2$): Theory ($R_1$) and Application ($R_2$). We impose a dependency constraint such that application-based learning requires sufficient theoretical background, modeled as:
\begin{equation}
g(\sum_{i=1}^{j} R_{1,i}) = a\cdot \max\{ (\sum_{i=1}^{j} R_{1,i}) - b, 0\},
\end{equation}
where $a$ is a scaling factor and $b$ represents the minimum theoretical foundation required before allocating applications.

\CoTutor{} generates optimal allocation plans under different objective functions: maximizing sentiment for individual students and optimizing for all students collectively (by maximizing the minimum sentiment across students). In the following, we consider three different types of $f(\cdot)$ and their solution. 

We first assume a linear model for $f(\cdot)$, defined as a linear combination of $R_{j,i}$ given by 
\begin{equation}
s_i = \theta s_{i-1} + (1 - \theta) \sum_{j=1}^{m} w_j R_{ji}.
\end{equation}
where the weight $w_i$ for the corresponding resources $R_{j,i}$ represents its attractiveness to the student. 

To define a meaningful $w_i$, we incorporate the assessment outcomes from the BKT model and the idea from educational psychology to describe the formulation of $f$. Since a moderate difficulty (or ``just‐right challenge'') fosters student engagement most, we assume an inverted-U shape relation between the assessment outcomes and the learning sentiment. Thus, $w_i$ is defined as follows
\begin{equation}
w_j = -\mathfrak{a}_j\big( P(L_{j,i})-\mathfrak{c_j}\big)^2+\mathfrak{b}_j,
\end{equation}
where $P(L_{j,i})\in [0,1]$ is the student's probability of having mastered activity $j$ at session $i$, $\mathfrak{c_j}\in (0,1)$ is the maximum level for activity $j$, i.e., the value of $P(L_{i,j})$ at which engagement or sentiment contribution is highest. $\mathfrak{b}_j$ is the maximum (peak) height of the function when $P(L_{i,j})=\mathfrak{c_j}$. Lastly, $\mathfrak{a}_j$ is a scaling factor controlling how sharply the sentiment falls off if the student's mastery is far below or above $\mathfrak{c_j}$. This parabola is the largest at $P(L_{i,j})=\mathfrak{c_j}$ and decreases symmetrically as the student's mastery deviates from $\mathfrak{c_j}$. This captures the \emph{optimal challenge principle}: if the student's skill in activity $j$ is much lower or higher than $\mathfrak{c_j}$, engagement drops (due to frustration or boredom). Figure \ref{fig:allocation_linear} illustrates the resource allocation outcomes for the linear model, showcasing five representative student profiles with distinct parameter configurations.

To incorporate nonlinear effects and better capture varying student responses, we introduce a power function model:
\begin{equation}
s_i = \theta s_{i-1} + (1 - \theta) \sum_{j=1}^{m} w_j R_{ji}^k,\quad 0 \leq k \leq 1. 
\end{equation}
As illustrated in Figure \ref{fig:allocation_power}, the power function model produces more flexible allocation trajectories, particularly benefiting students with strong resource dependencies. This formulation better aligns with knowledge-tracing principles and offers greater adaptability in strategy planning. Overall, the results demonstrate that \CoTutor{} effectively optimizes resource distribution, ensuring smooth and structured progression for students across different learning preferences. The flexibility in modeling various influence functions highlights its robustness in adapting to diverse educational settings.

\section*{Evaluation}\label{sec:evaluation}
To validate the efficacy of the proposed AI-assisted educational model, \CoTutor{}, we conducted a two-part evaluation through: a) publicly available educational datasets and b) a controlled experiment in a university-level course. The former focuses on model performance in tracking students' learning progress and predicting learning events, while the latter examines the impact of \CoTutor{} on student learning outcomes and engagement in a real-world classroom setting. In our evaluation of \CoTutor{}, we benchmarked its performance against three established models:
\begin{itemize}
    \item \textbf{BKT:} Foundational probabilistic model that estimates a student's mastery of skills over time.
    \item \textbf{DKT:} Advanced model employing recurrent neural networks to capture complex patterns in student learning sequences.
    \item \textbf{Code-based Deep Knowledge Tracing (Code-DKT):} A specialized model designed for programming education, which utilizes an attention mechanism to extract and select domain-specific code features, enhancing the traditional DKT framework.
\end{itemize}

By comparing \CoTutor{} with these models, we aim to comprehensively assess its effectiveness in modeling student learning processes, predicting future performance, and assisting personalized learning experiences.

\begin{table}[t]
\centering
\caption{Performance Comparison of \CoTutor{} and Baseline Models on EdNet Dataset}
\label{tab:benchmark_results}
\begin{tabular}{lccccc}
\toprule
\textbf{Model} & \textbf{Accuracy $\uparrow$} & \textbf{AUC-ROC $\uparrow$} & \textbf{PR-AUC $\uparrow$} & \textbf{RMSE $\downarrow$} & \textbf{NLL $\downarrow$}  \\
\midrule
BKT & 72.5\% & 0.690 & 0.682 & 0.256 & 0.671\\
DKT & 78.9\% & 0.714 & 0.725 & 0.241 & 0.625  \\
Code-DKT & 78.2\% & 0.705 & 0.740 & 0.247 & 0.618  \\
\textbf{\CoTutor{}} & \textbf{84.3\%} & \textbf{0.717} & \textbf{0.741} & \textbf{0.229} & \textbf{0.594} \\
\bottomrule
\end{tabular}
\end{table}

\subsection*{Benchmarking on Public Educational Datasets}

EdNet \cite{ednet} is a large-scale dataset collected from Santa, an Intelligent Tutoring System designed for TOEIC (Test of English for International Communication) preparation. The dataset comprises over 131 million interactions from more than 784,000 students over a period of two years. Students engage with Santa via multiple platforms (Android, iOS, and Web), interacting with exercises, video lectures, and expert commentaries. EdNet provides four subsets of data, ranging from the least to the most granular levels of student interaction. For our experiments, we focus on KT1, the subset that captures student responses at the interaction level, including timestamps, user IDs, item IDs, user answers, and skill labels. Additionally, we use metadata that provides the correct answers and knowledge component labels, which are critical for building a Q-matrix to map items to skills. 

To ensure high-quality data for model training and evaluation, we apply the following preprocessing steps: 1) Remove duplicate entries and interactions with missing user answers; 2) Filter out items with undefined skill tags, as they cannot be mapped to any known knowledge components; 3) Construct a binary q-matrix from the metadata dataset, ensuring each item maps correctly to its respective skills. 4) Remove students with fewer than 10 interactions, as insufficient data limits meaningful learning trajectory analysis.

The performance was assessed using the following metrics, which align with established practices in EdTech evaluation \cite{edtech_metrics}:
\begin{itemize}
    \item \textbf{Accuracy}: a common baseline for model comparison in knowledge tracing 
    \item \textbf{AUC-ROC}: for mastery classification due to its threshold-independence
    \item \textbf{PR-AUC}: better evaluate rare-event detection in imbalanced educational data
    \item \textbf{RMSE}: measures prediction error magnitude for probabilistic knowledge tracing models
    \item \textbf{NLL}: assesses calibration quality, critical for adaptive systems relying on confidence estimates
\end{itemize}

This combination follows established knowledge tracing research, where both discriminative ability and reliable confidence estimates are essential for adaptive learning systems. We prioritize these over metrics like Cohen's kappa, which are better suited for human-rater agreement studies.

Table \ref{tab:benchmark_results} summarizes the results of the benchmark evaluation, where \CoTutor{} outperforms the baseline models across all evaluation metrics, demonstrating its superior ability to track student progress and predict future learning events, thus supporting \textbf{H1}. The incorporation of signal processing techniques enhances real-time adaptability, allowing more precise modeling of student knowledge states.

\begin{table}[t]
\centering
\caption{Comparison of student assignment scores, prediction accuracy, and engagement metrics across the five experimental groups. Values in parentheses indicate statistics relative to the \CoTutor{} group ($n = 40$), shown as (t\textbar{}p\textbar{}d), where \textit{t} is the t-statistic, \textit{p} is the p-value, and \textit{d} is Cohen's d.}
\label{tab:sub_exp_results_tpd}
\begin{tabular}{lcccc}
\toprule
\textbf{Model} & \textbf{Avg Assignment Score} & \textbf{Post-Course Survey Score} & \textbf{Activity Participation Rate} & \textbf{Prediction Accuracy} \\
\midrule
Control & 68.4\% (2.78\textbar{}0.01\textbar{}0.62) & 67.9\% (4.71\textbar{}0.00\textbar{}1.05) & 64.2\% (3.41\textbar{}0.00\textbar{}0.76) & / \\
BKT & 70.1\% (1.79\textbar{}0.08\textbar{}0.40) & 69.4\% (3.27\textbar{}0.00\textbar{}0.73) & 67.7\% (2.18\textbar{}0.03\textbar{}0.49) & 59.0\% \\
DKT & 70.4\% (2.93\textbar{}0.00\textbar{}0.65) & 67.7\% (4.90\textbar{}0.00\textbar{}1.10) & 67.7\% (2.18\textbar{}0.03\textbar{}0.49) & 62.5\% \\
Code-DKT & 71.8\% (1.68\textbar{}0.05\textbar{}0.38) & 69.8\% (2.88\textbar{}0.01\textbar{}0.64) & 72.8\% (0.66\textbar{}0.51\textbar{}0.22) & 65.1\% \\
\textbf{\CoTutor{}} & 73.2\%(/) & 72.8\% (/) & 73.9\% (/) & 67.4\% \\
\bottomrule
\end{tabular}
\end{table}

\subsection*{Controlled Study in a University Course}
To assess the effectiveness of \CoTutor{} in predicting student performance from captured signals from known information resources and providing corresponding recommendations, we conducted a controlled study in a university AI course with 203 students, over a period of one semester. Students were randomly assigned to four experimental groups (each using one of the four models in Table \ref{tab:benchmark_results} for AI-assisted learning, with $n = 40-41$) and a control group ($n = 40$) that followed traditional methods. The module's pass/fail structure (historically $\geq$95\% pass rate) and attendance-based passing criteria (85\% threshold) ensured no academic disadvantage from group assignment.

Throughout the semester, students completed five graded assignments (A1-A5) along with lectures, tutorials, and other educational activities. They also participated in a post-course survey that collected feedback on the modules. Their participation rates in class activities (such as quizzes) were recorded.

Table~\ref{tab:sub_exp_results_tpd} summarizes the statistics for each of the five groups, including average assignment score, average post-course survey score, average participation rate in class activities, and model prediction accuracy of student performance. We performed t-tests comparing \CoTutor{} with other experimental groups (with $N_\mathrm{test} = N_\mathrm{control} = 40$, $\alpha = 0.05$, yielding critical t-values of 1.68 for one-tailed tests), reporting the t-value, p-value, and Cohen's d in brackets. The t-values for the \CoTutor{} group versus the baseline groups all exceed the critical thresholds for both $t$ and $p$, with medium to strong effect sizes based on Cohen's d, except for the Activity Participation Rate metric when compared to Code-DKT, which shows a borderline p-value and small effect size. In spite of this exception, the results provide directional insights into performance improvements, which we plan to validate in future work with a larger and more diverse student population.

\CoTutor{} has achieved the highest prediction accuracy (67.4\%), outperforming BKT (59.0\%), DKT (62.5\%), and Code-DKT (65.1\%). These results demonstrate \CoTutor{}'s effectiveness in integrating multimodal learning signals to accurately model and predict student performance, thereby supporting \textbf{H1}.

The experiment group with \CoTutor{} also achieved the highest average assignment score (73.2\%) across the five experiment groups, followed by Code-DKT (71.8\%), DKT (70.4\%), BKT (70.1\%), and the control group (68.4\%). These empirical findings align with \textbf{H2}, suggesting that \CoTutor{} is associated with improved learning outcomes compared to all baseline conditions. One potential contributing factor to this trend is that \CoTutor{}'s adaptive resource allocation and personalized intervention strategies—which dynamically adjust based on real-time student performance data—may help students focus on areas of weakness, potentially leading to better retention and understanding. While this interpretation remains speculative, it offers a plausible explanation for the observed performance differences and provides a foundation for further controlled studies.

The results also show outstanding outcomes for the four Experiment Groups across student engagement metrics. Specifically, the higher satisfaction from the post-course survey scores indicates that students found the AI-assisted learning experience more engaging and effective; the greater improvement over time highlights the system's ability to sustain and enhance learning outcomes; the higher participation rates suggest higher levels of engagement and motivation among students with learning journey assisted with knowledge tracing and generative AI techniques.

The benchmark and controlled study collectively validate the effectiveness of \CoTutor{} as an AI-driven educational framework. The benchmark results highlight its superior accuracy in modeling students' knowledge states and learning progression, while the user study demonstrates measurable improvements in learning outcomes. The model outperforms baseline methods in predictive accuracy and enhances student performance in real-world settings. The integration of signal processing techniques for real-time knowledge tracing enables more dynamic and personalized interventions, making \CoTutor{} an effective tool for AI-assisted learning. The higher engagement levels reported by students using \CoTutor{} suggest that personalized, AI-driven interventions can foster motivation and active participation, which are key drivers of academic success.

\subsection*{Discussion}
The evaluation of the \CoTutor{} model demonstrates its potential to revolutionize AI-assisted education in Computer Science and beyond. By maximizing key factors such as personalization, engagement, and scalability while minimizing ethical risks, the model provides a robust framework for enhancing learning outcomes. The results underscore the importance of iterative optimization, real-time feedback, and ethical considerations in designing AI-driven educational systems. As we continue to refine and expand this model, we aim to make high-quality, personalized education accessible to learners worldwide.

To highlight the unique contributions of \CoTutor{}, we compare it with other existing models to verify its advantages in personalization, scalability, and adaptability. Compared to predefined probability models and deep learning-based methods, \CoTutor{} dynamically adjusts resource allocation based on student progress, and it provides a more structured yet flexible optimization approach that increases interpretability while delivering superior performance, e.g., 7.8\% to 16.3\% increase in accuracy, as shown by Table \ref{tab:benchmark_results}. \CoTutor{}'s scalability is enhanced by its localized pre-LLM computation and caching mechanis,m which reduces LLM prompts by 30\%, yielding estimated cost savings of 15\% for GPT-4o (\$2.50 vs \$1.25 per million tokens) and 42\% for Claude 3.7 (\$3.75 vs \$0.3 per million tokens). Additionally, \CoTutor{} balances complexity and interpretability, with moderate implementation complexity making it suitable for real-world educational platforms without imposing excessive computational overhead. These advantages position \CoTutor{} as a deployable, ethically grounded solution for scalable AI-assisted education.

\section*{Ethical, Legal, and Societal Implications}
\CoTutor{} implements rigorous data protection measures to safeguard student privacy while enabling personalized learning support. The system processes only digitally recorded interactions from educational platforms, including:
\begin{itemize}
\item Assignment submissions (e.g., code) and associated metadata (timestamps, attempts),
\item Quiz/exam responses and grading logs,
\item Participation metrics (forum posts, video-watching patterns),
\item AI chatbot queries (Q\&A logs), and
\item Post-module feedback surveys.
\end{itemize}
Data is pseudonymized (with identifiers decoupled from direct user attribution) and encrypted during transmission and storage. Strict role-based access controls and differential privacy techniques ensure that personalized insights are accessible only to authorized educators, while raw individual performance data remains protected.

As an AI-driven system for personalized learning support, \CoTutor{} currently aligns with the high-risk category defined in the European Union (EU) AI Act (Annex III, Section 3(d)), as it dynamically adapts feedback and instructional strategies based on student progress. However, it is important to note that \CoTutor{} is not designed for determining access to educational programs (3(b)) or for grading or evaluating learning outcomes (3(c)). Future extensions—such as integration with admissions pipelines or automated assessment tools—could expand its applicability to those additional categories, and would require corresponding legal and procedural safeguards, including formal impact assessments by key privacy and data protection laws, technical documentation for regulatory review, and procedural controls (e.g., human-in-the-loop validation).

We acknowledge that high-risk classification under the EU AI Act entails specific requirements for transparency, data governance, human oversight, and post-deployment monitoring. While \CoTutor{} is currently presented as a research methodology evaluated within a controlled study population, any future production deployment would need to undergo formal risk assessment and compliance review aligned with applicable regional regulations—for example, complying with relevant data governance and accountability frameworks—addressing the legal, operational, and infrastructural components necessary for a safe and responsible rollout at scale.

\CoTutor{} integrates human oversight as both a regulatory requirement and pedagogical safeguard. The system provides adaptive suggestions while requiring explicit educator approval for consequential decisions, implemented through: (1) mandatory override justifications, (2) pre-deployment AI evaluation training, and (3) continuous educator feedback channels. 
This approach preserves instructor authority while leveraging AI's analytical strengths. It is important to develop automated monitoring tools to further streamline this AI-educator partnership, with the following real-time collaboration, auditing, and intelligent suggestion refinement:
\begin{itemize}
\item Immutable logs of AI suggestions, educator actions, and student outcomes,
\item Model cards documenting \CoTutor{}'s decision boundaries and confidence thresholds,
\item Quarterly transparency disclosures to institutional stakeholders.
\end{itemize}

\section*{Directions for Future Research Work}
Building on the foundation of our proposed model, which integrates key aspects of intelligent tutoring systems to enhance programming education, several promising directions can be explored to further advance AI-assisted education:
\begin{itemize}
\item \textbf{Hybrid Models with Multimodal Data}
Combining AI-driven insights with human expertise and integrating multimodal data (e.g., video, audio, interaction logs) can deepen personalization, improve engagement, and yield richer insights into student learning patterns.
\item \textbf{Privacy-Enhancing Technologies (PETs)}  
Future work will benchmark PETs such as federated learning, differential privacy, and hybrid encryption to protect student data while maintaining model performance, complementing existing safeguards like anonymization and access controls.  
\item \textbf{Supporting Educators}  
Extending the model to support curriculum design, instructional strategies, and performance monitoring can empower educators with data-informed tools. Research on collaborative learning, peer assessment, and student–teacher trust dynamics can further enhance pedagogical impact.  
\item \textbf{Extending to Other Disciplines}  
Adapting the framework beyond programming to other scientific domains by developing evaluation methods for diverse AI outputs would broaden applicability and educational value.  
\end{itemize}
By pursuing these directions, the proposed work could establish a transformative {\it human-AI cotutoring} paradigm designed to cultivate `{\it learning to learn}' skills, provide scalable educator support, and address the evolving challenges of modern education.

\section*{Conclusion}
In conclusion, we present \CoTutor{}, a comprehensive and ethically responsible AI-driven model for programming education, broadly applicable to any domain requiring coding skills. By framing an optimization problem that maximizes learning outcomes while mitigating risks, \CoTutor{} delivers personalized instruction, real-time feedback, and adaptive knowledge tracing through AI copilots. Its iterative, data-driven design fosters a responsive environment that adapts to student needs while nudging them to learn how to learn. \CoTutor{} empowers the next generation of programmers to leverage generative AI not as a crutch, but as a catalyst—enhancing creativity, initiative, and problem-solving. Just as Paul Buchheit built Google Gmail's first prototype in a single day without AI, future programmers should think boldly, iterate relentlessly, and turn ideas into impact—now supercharged with AI copilots. Future work will enhance interpretability, address privacy concerns, and expand real-world deployment toward a robust and inclusive AI-powered educational ecosystem.

%
\section*{Acknowledgment}
This work was supported in part by the Singapore Ministry of Education Academic Research Fund Tier 2 MOE-T2EP20224-0009 and the National Science and Technology Council, R.O.C., under Grant 114-2115-M-033-003-MY2.

%
\bibliographystyle{IEEEtran}
\bibliography{submission/references}

\section*{Biographies}
\label{sec:bio}
\vspace{-11pt}

\begin{IEEEbiographynophoto}{Yuchen Wang}
(~\IEEEmembership{Member,~IEEE}) is a Ph.D. candidate at Nanyang Technological University and a full-stack software engineer specializing in mobile development. She received her B.E. and M.S. degrees in Computer Science, both with Honours with the Highest Distinction, from the National University of Singapore. Her research mainly focuses on AI-Assisted Programming and Sound and Music Computing. Contact her at yuchen011@e.ntu.edu.sg. 
\end{IEEEbiographynophoto}
\vspace{-11pt} 

\begin{IEEEbiographynophoto}{Pei-Duo Yu}
(~\IEEEmembership{Member,~IEEE}) received the B.Sc. and M.Sc. degrees in applied mathematics from National Chiao Tung University, Taiwan, in 2011 and 2014, respectively, and the Ph.D. degree from the Department of Computer Science, City University of Hong Kong, Hong Kong. Currently, he is an Assistant Professor with the Department of Applied Mathematics. His research interests include combinatorics counting, graph algorithms, optimization theory, and its applications. Contact him at peiduoyu@cycu.edu.tw. 
\end{IEEEbiographynophoto}
\vspace{-11pt} 

\begin{IEEEbiographynophoto}{Chee Wei Tan}
(~\IEEEmembership{Senior Member,~IEEE}) received an M.A. and Ph.D. in Electrical Engineering from Princeton University. He is with Nanyang Technological University. He conducts research in networks, distributed optimization, and generative AI. Dr. Tan has served as the Editor for several journals, including IEEE Transactions on Signal and Information Processing, IEEE Transactions on Cognitive Communications and Networking, IEEE/ACM Transactions on Networking, IEEE Transactions on Communications and as IEEE ComSoc Distinguished Lecturer. He received the Princeton University Wu Prize for Excellence, Google Faculty Award, 2024 IEEE CAI Honorable Mention Award in Foundation Models and Generative AI, City University of Hong Kong Teaching Excellence Awards, and was selected twice for the U.S. National Academy of Engineering China-America Frontiers of Engineering Symposium. Contact him at cheewei.tan@ntu.edu.sg.
\end{IEEEbiographynophoto}

\vfill

\end{document}